\documentclass[letterpaper]{article}
\usepackage[preprint]{aaai2027}
\usepackage[hyphens]{url}
\usepackage{graphicx}
\usepackage{natbib}
\usepackage{caption}
\usepackage{booktabs}
\usepackage{amsmath}
\usepackage{amssymb}
\usepackage{multirow}
\usepackage{amsmath}
\usepackage{amssymb}
\usepackage{algorithm}
\usepackage{algpseudocode}
\usepackage{pifont}
\nocopyright

\newcommand{\method}{HiMA-MDD}

\newcommand{\hierarchicalgap}{hierarchical measurement--coordination gap}

\title{\includegraphics[width=1cm]{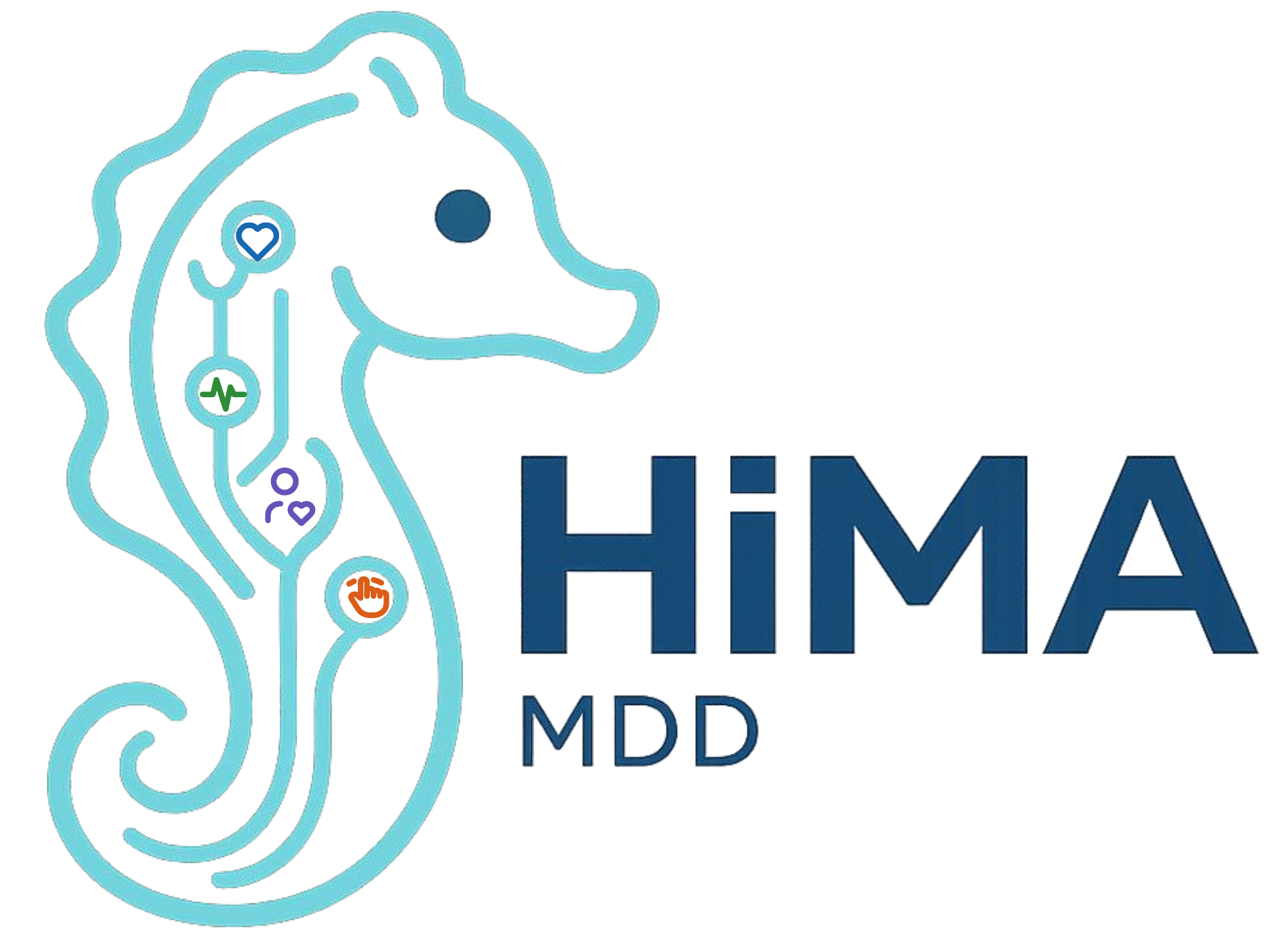}HiMA-MDD: A Hierarchical Multi-Agent Harness for Interpretable Multimodal Depression Detection in Clinical Interviews}
\author{Ao Chen, ~Xiaojiang Peng\corresponding}
\affiliations{School of Artificial Intelligence, Shenzhen Technology University}
\begin{document}

\maketitle

\begin{abstract}

Depression assessment from multimodal clinical interviews requires integrating dispersed evidence from multiple symptoms into a coherent PHQ‑8 profile. This process is hierarchical: relevant evidence is often sparse and context‑dependent within local question–answer exchanges, multiple exchanges jointly support symptom‑level judgments, and the final assessment depends on the coherence of the complete symptom profile. Existing LLM systems either process interviews holistically or distribute work across generic agent roles; neither design necessarily provides an explicit orchestration mechanism that coordinates evidence access, item-score authority, bounded feedback, and state recording across these levels. 
To address this gap, we introduce HiMA-MDD, a hierarchical multi-agent harness that aligns this assessment hierarchy with three agent layers. After non-agentic preprocessing constructs context-preserving multimodal QA units, Layer 1 identifies candidate QA-to-item relations and supports bounded item-grounded evidence routing. Layer 2 assigns symptom groups to operational factor specialists, with one specialist responsible for each provisional item score. Layer 3 audits the complete provisional profile, requests at most one round of targeted revision, and reconstructs the verified PHQ-8 profile. This layered design naturally yields a Hierarchical Evidence Trace, preserves all intermediate evidence, judgments, and revisions for auditability. The final item scores then deterministically produce the total score and screening decision. Using Qwen2.5‑72B‑Instruct as the harness backbone, our experiments on E‑DAIC demonstrate that HiMA‑MDD outperforms the compared state‑of‑the‑art methods.
\end{abstract}

\section{Introduction}
Depression is a common mental disorder characterized by persistent depressed mood or loss of interest and pleasure, often accompanied by changes in sleep, appetite, energy, concentration, and self-worth. It can disrupt relationships, education, employment, and everyday functioning, and severe episodes may be associated with suicide. The World Health Organization estimates that approximately 332 million people worldwide experience depression, including 5.7\% of adults, while substantial gaps in access to mental-health care remain~\cite{who2025depression}. Reliable and timely assessment is therefore an important clinical and public-health problem, motivating computational methods that can organize information dispersed throughout clinical interviews.

\begin{figure}[t]
    \centering
    \includegraphics[width=\columnwidth]{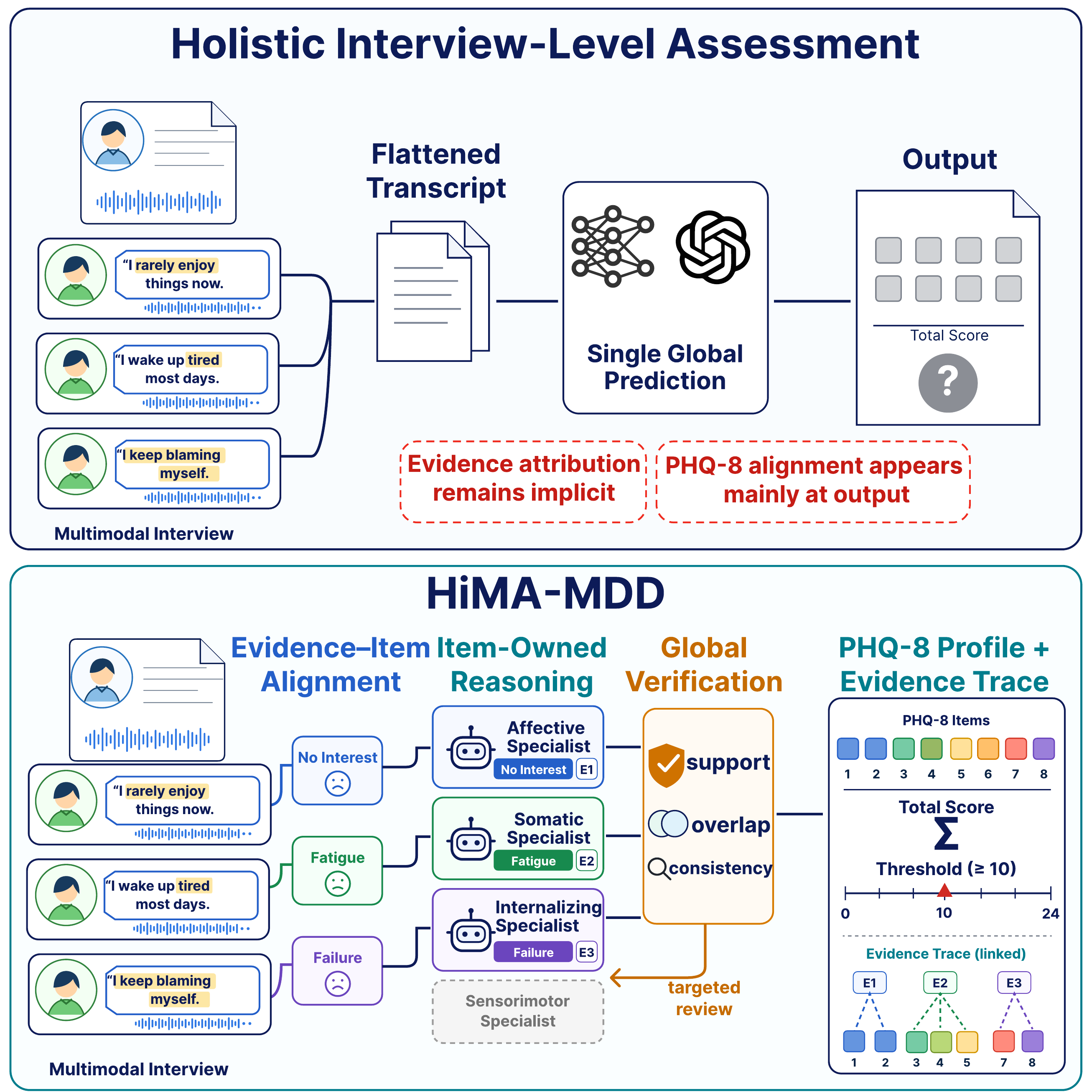}
    \caption{Comparison between holistic interview-level assessment and our HiMA-MDD.}
    \label{fig:motivation}
\end{figure}
Computational approaches to depression assessment include holistic prediction over complete interviews as well as methods that incorporate questionnaire structure, retrieve participant-specific evidence, organize long interview content, or produce symptom-aware outputs~\cite{nguyen2022questionnaires,jung2024hique,mandal2025questmf,wang2026mtsp,rosenman2024questionnaire,zhang2025red,chen2024sega,lyu2026depllm}. More recently, multi-agent systems have divided questioning, response-adequacy assessment, scoring, judgment, and updating among specialized functional roles~\cite{kim2024mdagents,hu2026agentmental,bi2025magi,greene2026psychiatrist}. Despite this progress, two related limitations remain. Holistic predictors may leave the connection between local interview context and symptom-level decisions implicit, whereas function-oriented multi-agent pipelines do not necessarily specify which evidence each agent may access or how scoring responsibility should align with the PHQ-8 item structure. More broadly, multimodal clinical interviews, symptom reasoning, and PHQ-8 assessment operate at different levels of granularity, but existing systems do not necessarily make the coordination of evidence access, scoring authority, and global revision explicit across these levels. We refer to this disconnect as the \hierarchicalgap{}. 
As illustrated in Figure~\ref{fig:motivation}, the schematic holistic pathway flattens the interview into a single global prediction, leaving evidence attribution implicit and applying PHQ-8 structure mainly at the output. 

To bridge this gap, we introduce \method{}, a measurement-aligned hierarchical multi-agent harness for interpretable depression assessment from multimodal clinical interviews, which maintains evidence--item responsibility from local multimodal exchanges to the final PHQ-8 profile. The harness is an explicit orchestration layer that governs what evidence each agent may access, which item scores it is authorized to produce or revise, how feedback is propagated across levels, and how intermediate states are preserved for auditing. Rather than treating multiple agents as an unconstrained workflow, \method{} aligns the evidence, agent, and measurement hierarchies through three governed agent roles. The QA-to-Item Grounding Agent identifies candidate relations between local interview exchanges and PHQ-8 items and supports the construction of bounded item-grounded evidence bundles. The Multi-Factor Symptom Reasoning Agent coordinates operational symptom-factor specialists, with one specialist responsible for each provisional item score. The Global Symptom Verification Agent audits the complete provisional profile, requests targeted revisions when needed, and reconstructs the verified eight-item PHQ-8 symptom profile. The final item scores are then summed to obtain the PHQ-8 total score and mapped to a screening decision under the prespecified threshold. Together, the recorded evidence links, intermediate judgments, audit findings, and revisions form a Hierarchical Evidence Trace that makes the evidence-to-profile process inspectable.

Our contributions are threefold:
\begin{itemize}
    \item \textbf{A hierarchical formulation of multimodal depression assessment,} connecting local interview evidence, symptom-level judgments, and the complete PHQ-8 profile to identify the \hierarchicalgap{}.
    \item \textbf{A measurement-aligned hierarchical multi-agent harness} that governs evidence access, single-owner provisional scoring, bounded feedback, profile reconstruction, and recorded provenance.
    \item \textbf{A layer-wise evaluation of harnessed reasoning} covering overall performance, evidence access, reasoning granularity, global verification and reconstruction, and trace inspectability.
\end{itemize}

\section{Related Work}
\subsection{Multimodal Depression Assessment}

Early multimodal depression-assessment systems commonly predicted a total score or screening label from complete interviews~\cite{alhanai2018depression,ringeval2019avec}. Recent systems have incorporated LLM-derived transcript representations, facial-expression features, and multimodal large language models into depression recognition~\cite{sadeghi2024multimodal,zhang2026mllmdr}. Other studies have moved toward finer-grained assessment by using questionnaire items or subscores as prediction targets, completing standardized questionnaires from interview text, or retrieving evidence separately for individual questionnaire items~\cite{mandal2025questmf,wang2026mtsp,rosenman2024questionnaire,ravenda2025arag}. Related approaches organize long interviews through question hierarchies or structural graphs and generate PHQ-aware symptom summaries, evidence, or rationales~\cite{zhang2025red,jung2024hique,chen2024sega,zheng2025emdrc,lyu2026depllm}. These studies show the value of preserving interview structure and predicting beyond a single total score. However, symptom-related evidence is often treated as an input representation or retrieval result rather than as part of an explicit control interface that also governs downstream scoring responsibility.

\subsection{Multi-Agent Systems for Depression Assessment}

Multi-agent systems provide another route to structured mental-health assessment. MDAgents adapts the composition of clinical decision teams to task complexity; AgentMental assigns question generation, response-adequacy assessment, scoring, and information updating to different agents; MAGI coordinates specialized roles around a structured psychiatric interview; and AI Psychiatrist Assistant combines assessment, judging, scoring, and review agents~\cite{kim2024mdagents,hu2026agentmental,bi2025magi,greene2026psychiatrist}. These systems show how functional specialization can organize complex assessment workflows. \method{} addresses a different coordination problem: assessment from a completed interview requires explicit control over which evidence each specialist may access, which provisional item scores it may produce, how global feedback may trigger a bounded revision, and which intermediate states are retained. It therefore uses the PHQ-8 measurement process to govern agent responsibilities rather than treating role specialization alone as the organizing principle.

\subsection{Psychometric Structure of the PHQ}

Psychometric research distinguishes the summed severity score of a questionnaire from the symptom profile represented by its individual items. PHQ scores are commonly obtained by summing item responses, yet people with the same total can have different symptom profiles~\cite{fried2015symptompatterns,fried2015sumscores}. Research on the internal structure of the PHQ-9 has examined several alternatives. A one-factor model treats all items as indicators of general depression severity. Correlated two-factor models commonly distinguish cognitive/affective and somatic dimensions, whereas bifactor models retain a general factor alongside more specific dimensions~\cite{lamela2020phq9review,fischer2022phq9latent,chae2025phq9meta}. Other work has reported a four-group organization comprising Affective, Somatic, Internalizing, and Sensorimotor symptoms~\cite{tseng2024stablephq9}. These findings do not establish a single universally accepted structure: the supported organization can vary with the population, instrument, and modeling assumptions. \method{} therefore does not propose or validate a new PHQ-8 factor model. Instead, it adapts the four-group organization as an operational responsibility map; because PHQ-8 omits the suicidality item, the Internalizing specialist is responsible only for low self-worth. The one-group, two-group, four-group, and item-wise configurations are evaluated as alternative reasoning granularities rather than competing psychometric models.

\section{Method}

\subsection{Task Formulation and Harness Control Interface}
\label{sec:method-formulation}

We introduce HiMA-MDD, a measurement-aligned hierarchical multi-agent harness for interpretable PHQ-8 assessment from completed multimodal clinical interviews. The harness is an explicit orchestration and control layer that implements an execution contract rather than merely naming a sequence of modules: it governs which evidence each role may inspect, which role is authorized to produce each provisional item score, how global feedback may revise that score, and which state transitions must be retained. The agents operate through these interfaces, while the harness constrains their permitted evidence access, score updates, revisions, and outputs.

Given a completed interview $X$, the core hierarchical harness produces eight raw verified PHQ-8 item scores $\{\hat y_i\}_{i=1}^{8}$, the corresponding total score $\hat S$, a screening decision $\hat c$, and a Hierarchical Evidence Trace $T$:
\begin{equation}
\label{eq:task-output}
\begin{aligned}
f(X)&\rightarrow(\{\hat y_i\}_{i=1}^{8},\hat S,\hat c,T),\\
\hat S&=\sum_{i=1}^{8}\hat y_i,\qquad
\hat c=\mathbb{I}[\hat S\ge 10],
\end{aligned}
\end{equation}
where each $\hat y_i\in\{0,1,2,3\}$. Within the core harness, the total score and screening decision follow the fixed PHQ-8 rule. The complete HiMA-MDD assessment harness additionally exposes an optional supervised calibration interface, described below, which maps the frozen raw state to calibrated item scores and then recomputes the total score and screening decision under the same fixed PHQ-8 rule.

We formalize the HiMA-MDD Hierarchical Harness as
\begin{equation}
\mathcal H_{\mathrm{HiMA}}
=
(\mathcal I,\mathcal R,\Omega,\mathcal V,\Gamma,\mathcal T),
\end{equation}
where $\mathcal I$ contains the PHQ-8 item rubrics and ordinal scoring semantics; $\mathcal R$ defines candidate QA-to-item relations, preliminary evidence-polarity metadata, and the bounded evidence-access policy; $\Omega$ maps every item to exactly one provisional score owner; $\mathcal V$ specifies cross-factor audit, targeted revision, and verified-profile reconstruction; $\Gamma$ deterministically maps the raw verified item vector to its total score and screening decision; and $\mathcal T$ specifies the provenance and state transitions retained in $T$. These interfaces enforce four invariants. First, grounding estimates candidate item relevance and preliminary evidence polarity but does not determine symptom severity or produce item scores. Second, each item has exactly one role authorized to write its provisional score, although evidence may be relevant to multiple items. Third, global feedback is targeted, limited to at most one revision round, and recorded together with the resulting response and score change. Fourth, neither an agent nor the verifier independently generates the total score or screening label; both are deterministic consequences of the eight raw verified item scores.

\begin{figure*}[t]
    \centering
    \includegraphics[width=0.99\textwidth]{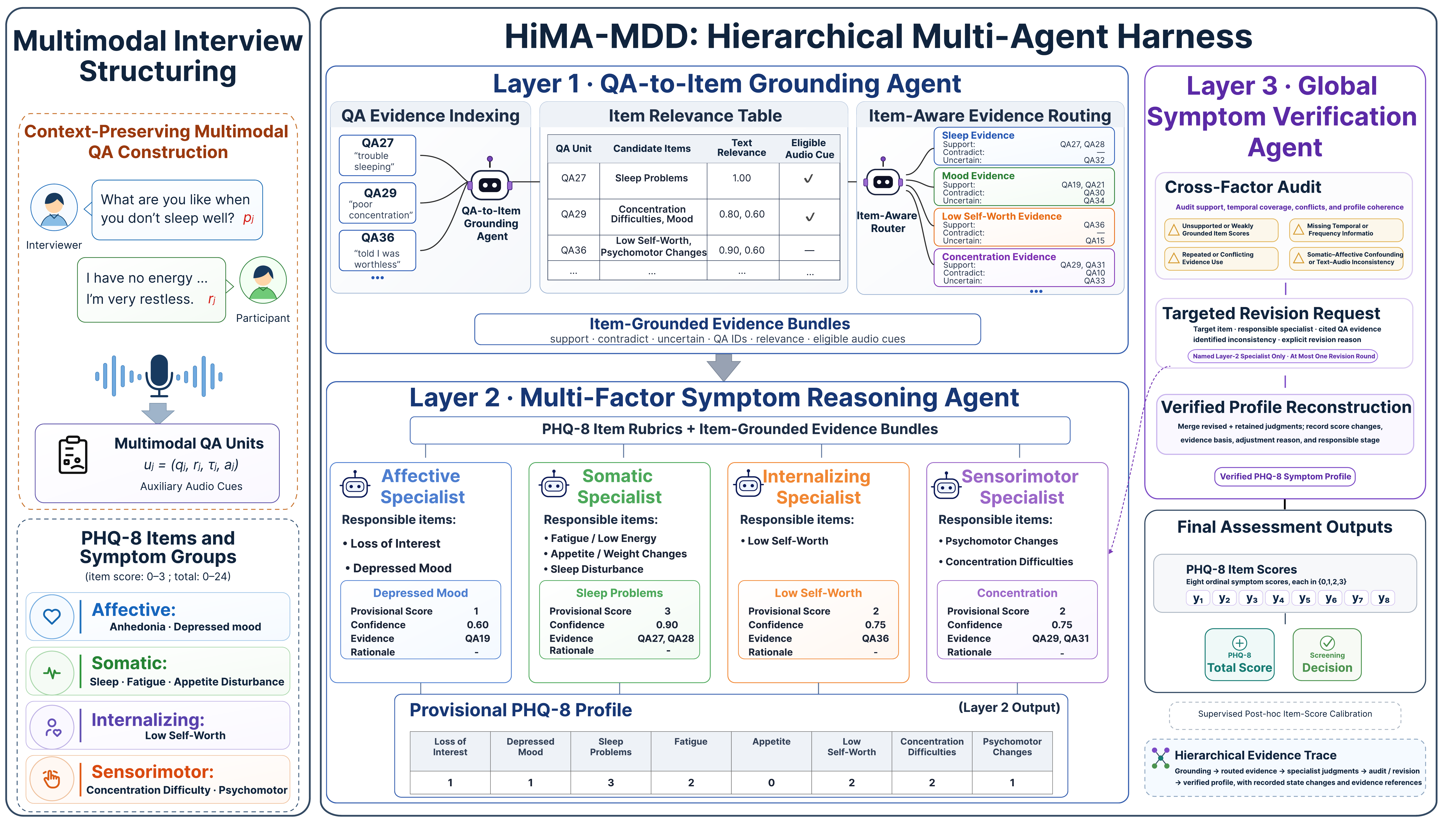}
    \caption{Overview of the HiMA-MDD assessment framework.}
    \label{fig:framework}
\end{figure*}

Figure~\ref{fig:framework} instantiates this contract from interview structuring to assessment output. Non-agentic preprocessing constructs context-preserving multimodal QA units and the PHQ-8 measurement contract. Layer~1 builds candidate QA-to-item relations and item-grounded evidence bundles under $\mathcal R$; Layer~2 assigns these bundles to single-owner factor specialists under $\Omega$ and produces the provisional profile; and Layer~3 applies the audit, targeted revision, and verified-profile reconstruction policy $\mathcal V$. The raw verified item vector is passed through $\Gamma$, while $\mathcal T$ retains the grounding, routed evidence, specialist judgments, audit and revision records, and verified profile as the Hierarchical Evidence Trace. The dashed Post-hoc Item-Score Calibration stage represents an optional supervised, measurement-constrained output-control interface in the complete HiMA-MDD assessment harness. It operates only after the core agentic execution is complete, without rerunning the agents or modifying the Hierarchical Evidence Trace, and recalibrates the eight item scores before the total score and screening decision are recomputed.

\subsection{Data and Measurement Structuring}
\label{sec:method-structuring}

HiMA-MDD first constructs context-preserving Multimodal QA Units using a question-based structure inspired by HiQuE~\cite{jung2024hique}. Each interviewer question is paired with all consecutive participant response turns before the next question. The resulting unit is
\begin{equation}
u_j=(q_j,r_j,\tau_j,a_j),\qquad j\in\{1,\ldots,J\},
\end{equation}
where $q_j$ is the interviewer question, $r_j$ is the grouped participant response, $\tau_j$ is the turn and timestamp metadata, and $a_j$ represents any aligned participant-speech acoustic descriptors. Retaining the interviewer question provides local context for short or elliptical responses, while the participant response remains the primary basis for symptom assessment~\cite{burdisso2024promptbias}.

The PHQ-8 measurement contract supplies eight item rubrics with $0$--$3$ ordinal scoring semantics over the preceding two weeks, the symptom groups that define specialist responsibility, and the fixed rule for computing the total score and screening decision~\cite{kroenke2009phq8}. These rubrics accompany the routed evidence throughout the reasoning layers. Participant-speech acoustic descriptors are aligned to QA units by timestamp. They combine clip-level eGeMAPS functionals extracted with the eGeMAPSv02 configuration~\cite{eyben2010opensmile,eyben2016gemaps} and depression and anxiety estimates computed over longer speech contexts by the KintsugiHealth Depression--Anxiety Model (DAM). The DAM model card reports training and evaluation on approximately 863 hours of speech from 35,000 individuals, collected via phone, tablet, or web app and labeled using clinician-administered or self-reported PHQ-9 and GAD-7; it does not list E-DAIC as a training or evaluation source~\cite{kintsugihealth2026dam}. The descriptors are verbalized and attached to audio-relevant QA evidence as auxiliary cues for the corresponding symptom judgments. Appendix~A describes transcript recovery, feature extraction, and descriptor verbalization.

\subsection{Layer 1: QA-to-Item Grounding}
\label{sec:method-grounding}

The QA-to-Item Grounding Agent converts the Multimodal QA Units into an item-level evidence index. For each unit, an LLM relevance mapper identifies candidate PHQ-8 items, estimates item-specific text relevance, assigns preliminary support, contradiction, or uncertainty metadata, records eligible audio cues, and provides a brief mapping rationale. These signals describe candidate relevance and evidence polarity; they do not determine symptom severity or produce item scores. Because a single exchange may inform several symptoms, the resulting Item Relevance Table supports many-to-many QA-to-item relations. An item-aware router then ranks the indexed units separately for each PHQ-8 item, applies lexical fallback when the initial mapping is empty, removes duplicate records, and limits the evidence supplied downstream. It organizes the selected records into supporting, contradictory, and uncertain evidence categories, producing an Item-Grounded Evidence Bundle for each item. Every record retains its QA identifier, question, response, timing, relevance and polarity metadata, and available audio cues. Layer~1 thereby determines which evidence each specialist can inspect; symptom severity assessment and item scoring begin in Layer~2.

\subsection{Layer 2: Multi-Factor Symptom Reasoning}
\label{sec:method-reasoning}

The Multi-Factor Symptom Reasoning Agent is instantiated by four parallel specialists. The Affective Specialist assesses Anhedonia and Depressed mood; the Somatic Specialist assesses Sleep disturbance, Fatigue, and Appetite disturbance; the Internalizing Specialist assesses Low self-worth; and the Sensorimotor Specialist assesses Concentration difficulty and Psychomotor disturbance. This organization is informed by prior analyses of PHQ symptom structure~\cite{tseng2024stablephq9,gunzler2020rdocphq9}. Each specialist receives the PHQ-8 rubrics and Item-Grounded Evidence Bundles for its assigned symptoms and returns provisional item scores, confidence estimates, cited supporting and contradictory evidence, evidence sufficiency, indications of missing frequency or temporal information, and concise judgments. Each item has one specialist responsible for its provisional score, while relevant QA evidence may be shared across items. The four specialist reports are combined into the Provisional PHQ-8 Profile supplied to Layer~3.

\subsection{Layer 3: Global Symptom Verification}
\label{sec:method-verification}

The Global Symptom Verification Agent receives the Provisional PHQ-8 Profile together with the specialist judgments, cited QA identifiers, and grounding and routing metadata, and applies a three-stage verification procedure. First, Cross-Factor Audit checks for unsupported scores, missing temporal or frequency information, repeated or conflicting evidence, somatic--affective confounding, and text--audio inconsistency; each issue is linked to the affected symptom and supporting QA references. Second, when reconsideration is warranted, a Targeted Revision Request specifies the symptom, responsible specialist, relevant QA identifiers, and issue to reconsider. Each implicated specialist receives its previous report, full assigned evidence bundle, and targeted audit instruction. Specialists that are not implicated are not rerun, whereas other items owned by a rerun specialist are requested to remain stable but may be regenerated. HiMA-MDD permits at most one revision round. Third, Verified Profile Reconstruction reconciles the revised and retained judgments with the audit findings, selected QA evidence, compact interview context, and PHQ-8 rubrics. A centralized LLM aggregator records score changes, addresses remaining cross-factor inconsistencies, and produces the Verified PHQ-8 Profile. The fixed PHQ-8 rule converts the raw verified item scores into $\hat S$ and $\hat c$. Throughout the hierarchy, the harness records QA-to-item relations, routed evidence bundles, specialist judgments, provisional scores, audit findings, revision requests and responses, score changes, and the verified profile. These records form the Hierarchical Evidence Trace, allowing each raw verified item score to be inspected alongside its routed evidence and recorded intermediate judgments.

\subsection{Post-hoc Item-Score Calibration}
\label{sec:method-calibration}

Post-hoc Item-Score Calibration implements the optional supervised output-control interface of the complete HiMA-MDD assessment harness. Motivated by systematic scoring biases that may arise when LLM judgments are mapped to ordinal scale items~\cite{zheng2023llmjudge,hada2024llmevaluators,jin2026llmdepressionscreening}, it converts the frozen raw HiMA-MDD state into eight calibrated item scores without rerunning the agent hierarchy or modifying the Hierarchical Evidence Trace. Its inputs are standardized numerical features summarizing the provisional and verified scores, score changes, confidence estimates, evidence retrieval, acoustic metadata, and audit and revision records. A separate Bayesian Ridge regressor is learned for each PHQ-8 item.

Each regressor is first fitted on the training partition, and its hyperparameters are selected on the development partition. The selected regressor is then refitted on the combined training and development data and used to calibrate the raw test output for that item. Test labels are used only for final evaluation. Each prediction is rounded to the nearest integer and clipped to the PHQ-8 range of $0$--$3$. The eight calibrated item scores are summed to obtain the final total score, and a total of at least 10 yields a positive screening decision. These calibrated item scores, the resulting total, and the screening decision constitute the complete-system outputs reported for E-DAIC, while the raw outputs are retained for evaluating the core agentic harness.

\section{Experiments}
\label{sec:experiments}

\paragraph{Dataset.}
We evaluate \method{} on E-DAIC and DAIC-WOZ, two multimodal clinical-interview corpora with PHQ-8 labels~\cite{gratch2014daic,ringeval2019avec}. DAIC-WOZ contains 189 Wizard-of-Oz interviews in which the virtual interviewer Ellie is controlled by a human interviewer. E-DAIC extends this corpus to 275 participants and adds interviews conducted by a fully autonomous AI interviewer; its test set consists entirely of autonomous interviews, providing a more difficult interview condition~\cite{gomezzaragoza2026autonomous}. Following established evaluation settings~\cite{wang2026mtsp,sadeghi2024multimodal,hu2026agentmental}, we use the held-out E-DAIC test split for the main evaluation and the fixed DAIC-WOZ development split for the additional robustness analysis. Both evaluations use ASR-derived transcripts and participant-speech audio. Each locally evaluated method predicts the eight PHQ-8 item scores on a $0$--$3$ scale; their sum gives the total score, and a total of at least 10 indicates a positive depression screen. Appendix~A provides preprocessing and target details.

\paragraph{Baselines.}
We compared \method{} with three prompt-based baselines (Zero-Shot, 3‑Shot, CoT~\cite{wei2022cot}) and two recent multi‑agent systems: MDAgents~\cite{kim2024mdagents} and AgentMental~\cite{hu2026agentmental}. We reran these baselines on the same E-DAIC test set using Qwen2.5-72B-Instruct~\cite{qwen2024qwen25}with the same PHQ‑8 rubric and output format, predicting eight item scores (summed to total and screening decision). For AgentMental, we replaced its original participant simulator (DeepSeek‑R1‑Distill‑Qwen‑32B) with Qwen2.5‑72B‑Instruct, feeding the completed interview transcript as the response source. Table~\ref{tab:overall} additionally includes source-reported E-DAIC results from the multimodal MLlm-DR~\cite{zhang2026mllmdr} and from the transcript-based Dep-LLM study, which reports Dep-LLM and several general LLMs~\cite{lyu2026depllm}. Appendix~B provides implementation and adaptation details.

\paragraph{Evaluation.}
The primary evaluation focuses on PHQ-8 total-score estimation and the thresholded screening decision. The protocol-matched comparison reports Total MAE and RMSE together with screening accuracy, $\kappa$, class-wise F1, and Macro-F1. These outcomes are complementary: total-score error measures aggregate severity estimation, whereas screening metrics test the decision induced by the predicted profile. Class-wise F1 reveals whether aggregate screening performance is dominated by one class. Appendix~B defines the metrics.

\begin{table}[t]
\centering
\tiny
\renewcommand{\arraystretch}{1.12}
\setlength{\tabcolsep}{1.0pt}
\newcommand{\NA}{\multicolumn{1}{c}{\textemdash}}
\begin{tabular*}{\columnwidth}{@{\extracolsep{\fill}}p{0.18\columnwidth}p{0.21\columnwidth}*{7}{c}@{}}
\toprule
\multicolumn{1}{c}{Group} &
\multicolumn{1}{l}{Method} &
\shortstack{MAE$\downarrow$} &
\shortstack{RMSE$\downarrow$} &
Acc.$\uparrow$ &
\shortstack{Screening\\$\kappa\uparrow$} &
F1[C]$\uparrow$ &
F1[D]$\uparrow$ &
\shortstack{Macro\\F1$\uparrow$} \\
\midrule
\multirow{6}{*}{\shortstack[l]{Reported\\Results}}
& MLlm-DR & 3.57 & 4.83 & \NA & \NA & \NA & \NA & 0.69 \\
& GPT-5.5 & \NA & \NA & 0.70 & \NA & 0.75 & 0.61 & 0.68 \\
& Gemini-3.1-Pro & \NA & \NA & 0.73 & \NA & 0.77 & 0.68 & 0.73 \\
& Claude-Opus-4.6 & \NA & \NA & 0.71 & \NA & 0.76 & 0.65 & 0.71 \\
& DeepSeek-V4 & \NA & \NA & 0.71 & \NA & 0.78 & 0.60 & 0.69 \\
& \shortstack[l]{Dep-LLM\\(Gemma3-12B-\\Instruct)} & \NA & \NA & 0.75 & \NA & 0.80 & 0.67 & 0.73 \\
\midrule
\multirow{5}{*}{\shortstack[l]{Re-implemented\\Baselines}}
& Zero-Shot & 4.54 & 6.13 & 0.73 & 0.35 & 0.82 & 0.48 & 0.65 \\
& 3-Shot & 4.39 & 5.98 & 0.73 & 0.36 & 0.81 & 0.52 & 0.67 \\
& CoT & 4.27 & 5.63 & 0.75 & 0.42 & 0.82 & 0.59 & 0.70 \\
& MDAgents & 3.98 & 5.22 & 0.75 & 0.45 & 0.81 & 0.63 & 0.72 \\
& AgentMental & 4.57 & 6.22 & 0.75 & 0.50 & 0.77 & 0.72 & 0.75 \\
\midrule
\multirow{2}{*}{HiMA-MDD}
& HiMA-MDD (raw) & 3.96 & $4.95^{*}$ & 0.80 & 0.57 & 0.85 & $0.72^{*}$ & 0.78 \\
& \shortstack[l]{\textbf{HiMA-MDD w/ }\\\textbf{Calibration}} & \textbf{3.41}$^{*}$ & \textbf{4.57}$^{*}$ & \textbf{0.84}$^{*}$ & \textbf{0.63} & \textbf{0.88}$^{*}$ & \textbf{0.74}$^{*}$ & \textbf{0.81} \\
\bottomrule
\end{tabular*}
\caption{Comparison of HiMA-MDD with baselines on E-DAIC.Results marked with * are better than Zero-Shot ($p<0.05$) based on metric-specific one-tailed paired tests.}
\label{tab:overall}
\end{table}

\paragraph{Implementation details.}

We implemented \method{} as a stateful LangGraph~\cite{langchain2026langgraph} workflow. All local systems used Qwen2.5‑72B‑Instruct (temperature 0), the same PHQ‑8 rubric, and shared evaluation code, with inference accelerated via vLLM~\cite{kwon2023vllm} on NVIDIA RTX 5880 Ada GPUs. For post-hoc calibration output, eight Bayesian Ridge regressors were fitted on the training set, hyperparameter-tuned on the development set, and applied to frozen raw \method{} test predictions. The screening threshold was fixed at 10 as common. An asterisk indicates an improvement over Zero-Shot at $p<0.05$; Appendix~C provides the statistical-testing details.


\subsection{Main Results}

\begin{table}[t]
\centering
\scriptsize
\setlength{\tabcolsep}{2.6pt}
\begin{tabular*}{\columnwidth}{@{\extracolsep{\fill}}lrrrr@{}}
\toprule
Method & \shortstack{Total\\MAE$\downarrow$} & Acc.$\uparrow$ & \shortstack{Screening\\$\kappa\uparrow$} & \shortstack{Macro\\F1$\uparrow$} \\
\midrule
Zero-Shot & 3.9143 & 0.7143 & 0.2081 & 0.5536 \\
3-Shot & 3.7714 & 0.7143 & 0.2457 & 0.5949 \\
CoT & 3.4286 & 0.7429 & 0.3046 & 0.6182 \\
MDAgents & \textbf{3.0571} & 0.8000 & 0.4842 & 0.7281 \\
AgentMental & \textbf{3.0571} & 0.7714 & 0.5122 & 0.7552 \\
HiMA-MDD (raw) & 3.9714 & \textbf{0.8571} & \textbf{0.6765} & \textbf{0.8381} \\
\bottomrule
\end{tabular*}
\caption{Results on the DAIC-WOZ development split. Bold indicates the best performance. Following the established DAIC-WOZ development-cohort protocol, all rows report raw outputs using ASR-derived transcripts.}
\label{tab:daic-dev}
\end{table}

Table~\ref{tab:overall} reports PHQ-8 total-score estimation and screening results on the held-out E-DAIC test split. HiMA-MDD + Post-hoc Calibration produces the final system output. It achieves a Total MAE of 3.4107 and a Macro-F1 of 0.8130, improving every reported metric over the raw harness output. These are also the best displayed Total MAE and Macro-F1 values among the source-reported and locally evaluated methods. The source-reported rows provide cross-paper context, while the locally rerun rows form the protocol-matched comparison using the same test split, PHQ-8 scoring protocol, and statistical analysis. Among the locally rerun baselines, CoT performs better than Zero-Shot and 3-Shot on Total MAE, Total RMSE, $\kappa$, F1[D], and Macro-F1. MDAgents further reduces total-score error, whereas AgentMental obtains the highest F1[D] and Macro-F1 among the baselines. The raw HiMA-MDD output supports mechanism analysis and uncalibrated comparison; among the uncalibrated local methods, it achieves the lowest Total MAE and RMSE and the highest accuracy, $\kappa$, F1[C], and Macro-F1, while AgentMental remains slightly higher on F1[D].

Table~\ref{tab:daic-dev} reports raw-system results on the DAIC-WOZ development split using the E-DAIC prompts and scoring configuration. HiMA-MDD (raw) achieves the highest accuracy, screening $\kappa$, and Macro-F1, increasing Macro-F1 from 0.7552 for the best-performing baseline to 0.8381. Its Total MAE is higher, showing that screening performance and absolute total-score error rank the methods differently. Overall, HiMA-MDD maintains robust screening performance on DAIC-WOZ with ASR-derived transcripts.

\begin{table}[t]

\centering
\scriptsize
\setlength{\tabcolsep}{1.8pt}
\begin{tabular*}{\columnwidth}{@{\extracolsep{\fill}}lrrrrr@{}}
\toprule
Variant & Acc.$\uparrow$ & $\kappa\uparrow$ & F1[C]$\uparrow$ & F1[D]$\uparrow$ & Macro-F1$\uparrow$ \\
\midrule
HiMA-MDD & \textbf{0.8036} & \textbf{0.5686} & \textbf{0.8493} & \textbf{0.7179} & \textbf{0.7836} \\
w/o \ding{172} & 0.7679 & 0.4902 & 0.8219 & 0.6667 & 0.7443 \\
w/o \ding{173}  & 0.7857 & 0.5248 & 0.8378 & 0.6842 & 0.7610 \\
w/o \ding{174} & 0.7857 & 0.5340 & 0.8333 & 0.7000 & 0.7667 \\
\bottomrule
\end{tabular*}
\caption{Component ablations of our HiMA-MDD. All rows report raw outputs w/o calibration. \ding{172} Cross-factor audit and targeted revision. \ding{173} Centralized verified-profile reconstruction. \ding{174} Acoustic descriptor augmentation.}
\label{tab:ablations}
\end{table}

\begin{figure*}[t]
\centering
\includegraphics[width=\textwidth]{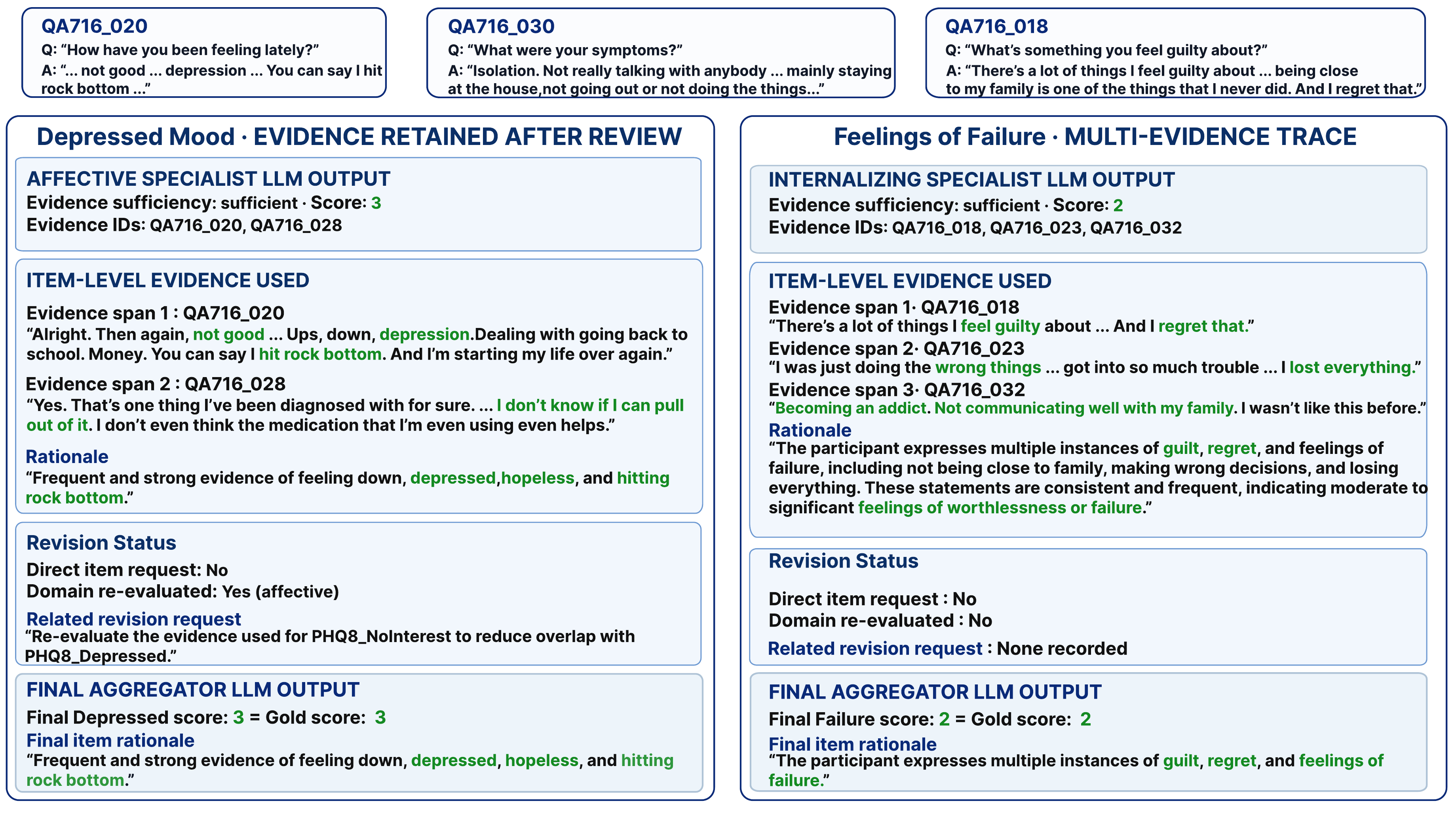}
\caption{Hierarchical Evidence Trace for E-DAIC participant 716, showing evidence retention after domain review for Depressed Mood and multi-evidence aggregation for Feelings of Failure.}
\label{fig:case_study}
\end{figure*}

\subsection{Effect of Reasoning Granularity and Evidence}

\begin{table}[t]
\centering
\scriptsize
\setlength{\tabcolsep}{2.4pt}
\begin{tabular*}{\columnwidth}{@{\extracolsep{\fill}}p{0.38\columnwidth}rrrr@{}}
\toprule
\shortstack[l]{Reasoning\\Configuration} & \shortstack{Total\\MAE$\downarrow$} & \shortstack{Total\\RMSE$\downarrow$} & \shortstack{Screening\\$\kappa\uparrow$} & \shortstack{Macro\\F1$\uparrow$} \\
\midrule
Single Agent & 4.3036 & 5.2627 & 0.4667 & 0.7333 \\
Two-Factor & 4.3571 & 5.3352 & \textbf{0.5000} & \textbf{0.7499} \\
Four-Factor (Default) & \textbf{4.0714} & \textbf{5.0533} & 0.4902 & 0.7443 \\
Item-Specific & 4.3393 & 5.3802 & 0.4563 & 0.7278 \\
\shortstack[l]{Four-Factor w/\\Shared Evidence} & 4.2500 & 5.0815 & 0.4340 & 0.7169 \\
\bottomrule
\end{tabular*}
\caption{Effect of reasoning granularity and evidence access. Configurations share candidate QA-to-item grounding, backbone, rubric, acoustic descriptors, and test cases, and the comparison ends after Layer~2 symptom reasoning.}
\label{tab:decomposition}
\end{table}

Table~\ref{tab:decomposition} examines how reasoning granularity and evidence access affect performance before global verification. The Single Agent, Two-Factor, Four-Factor, and Item-Specific configurations use one, two, four, and eight reasoning agents, respectively. The Single Agent receives a capped evidence bundle retrieved across all eight items; the factor- and item-based configurations receive bounded evidence for their assigned symptoms; and the shared-evidence condition gives the four Factor Specialists the same capped bundle. Two-Factor Specialists divide responsibility between a cognitive--affective group covering Anhedonia, Depressed mood, Low self-worth, Concentration difficulty, and Psychomotor disturbance, and a somatic group covering Sleep disturbance, Fatigue, and Appetite disturbance~\cite{patel2019phq9invariance,gunzler2020rdocphq9}. The four-factor responsibility map separates affective, somatic, internalizing, and sensorimotor responsibilities, following prior analyses of PHQ symptom structure~\cite{tseng2024stablephq9}. Item-Specific Specialists assign one reasoning agent to each PHQ-8 item, whereas the Single Agent condition scores all items together. The results reveal a granularity trade-off: Four-Factor Specialists obtain the lowest Total MAE and RMSE, Two-Factor Specialists achieve the highest $\kappa$ and Macro-F1, and Item-Specific Specialists score lower than both configurations on score estimation and screening. Thus, four specialists provide the strongest total-score estimation, whereas two specialists favor screening agreement. With the four-factor responsibility map fixed, the item-grounded bounded-access condition performs better than capped shared access on every reported metric. Because the two access conditions differ in both evidence composition and context length, the supported comparison is between the complete bounded-access and capped shared-access policies.

\subsection{Component Ablations of the HiMA-MDD Harness}

Table~\ref{tab:ablations} evaluates the contributions of cross-factor audit with targeted revision, centralized verified-profile reconstruction, and acoustic descriptor augmentation to the complete raw harness. All three ablations reduce every screening metric. Removing Cross-Factor Audit and Targeted Revision produces the largest decrease, showing that the audit--revision path contributes most strongly among the tested components. Removing centralized reconstruction also lowers performance, indicating that directly combining specialist outputs is less effective than jointly reconciling the verified profile. The decline without acoustic descriptors shows that participant-speech cues provide useful auxiliary information.

\subsection{Case Study}

Figure~\ref{fig:case_study} shows how the Hierarchical Evidence Trace connects item scores to distributed interview evidence and subsequent review decisions for participant 716. For Depressed Mood, the Affective Specialist assigns a score of 3 using QA716\_020 and QA716\_028, which describe depression, hopelessness, and hitting rock bottom. Although a related evidence-overlap request triggers re-evaluation of the affective domain, the item is not directly targeted and its evidence and score are retained; the final aggregator records the same score and rationale. For Feelings of Failure, the Internalizing Specialist combines three evidence spans concerning guilt, regret, harmful decisions, and family disconnection to support a score of 2. The trace records the responsible specialist, cited evidence spans, evidence sufficiency, review scope, and final item rationale, allowing both the retained judgment and the multi-evidence judgment to be inspected against their source responses.

\section{Conclusion}
In this paper, we have presented HiMA-MDD, a hierarchical multi-agent harness for PHQ-8 assessment from completed multimodal clinical interviews. HiMA-MDD has organized assessment across evidence grounding, factor-level symptom reasoning, and global symptom verification, producing item scores, total severity, screening decisions, and a recorded Hierarchical Evidence Trace. Results on E-DAIC support the utility of measurement-aligned governance for organizing PHQ-8 assessment while maintaining an inspectable path from interview evidence to the final symptom profile.

\clearpage
\bibliography{aaai2027}

\section{Technical Appendix}
\label{sec:technical-appendix}

\subsection{Appendix A: Dataset, Preprocessing, and Targets}

\subsubsection{E-DAIC}

The Extended Distress Analysis Interview Corpus (E-DAIC) was released with the AVEC 2019 Detecting Depression with AI challenge~\cite{ringeval2019avec}. It extends DAIC-WOZ with interviews conducted by a fully autonomous virtual interviewer. The corpus contains 275 participant sessions and provides audio, video-derived descriptors, automatic transcripts, and depression labels. The official test partition consists entirely of autonomous-agent sessions. HiMA-MDD uses processed transcripts and participant-speech audio.

\begin{table}[H]
\centering
\small
\begin{tabular}{@{}lrp{0.52\columnwidth}@{}}
\toprule
Partition & Participants & Role in this study \\
\midrule
Train & 163 & Post-hoc calibration fitting and final refit \\
Development & 56 & Post-hoc calibration selection and final refit; three-shot pool \\
Test & 56 & Final evaluation only \\
\bottomrule
\end{tabular}
\caption{Official E-DAIC partitions and their use in this study. The test partition is reserved for final reporting.}
\label{tab:appendix-edaic-splits}
\end{table}

\subsubsection{DAIC-WOZ}

The Distress Analysis Interview Corpus--Wizard of Oz (DAIC-WOZ) contains clinical interviews with 189 participants conducted by the virtual interviewer Ellie under a Wizard-of-Oz protocol, in which a hidden human interviewer controls the interaction~\cite{gratch2014daic}. The corpus provides interview transcripts, audio recordings, videos, and PHQ-8 labels. Following the development-cohort evaluation setting used in prior work~\cite{hu2026agentmental}, we evaluate the locally implemented methods on the fixed 35-participant development split. Our inputs are ASR-derived transcripts reconstructed from the corresponding E-DAIC audio and participant-speech audio rather than the official DAIC-WOZ transcripts. We retain the E-DAIC experimental configuration without DAIC-specific prompt modification or calibration. E-DAIC remains the primary dataset for held-out test evaluation; the DAIC-WOZ cohort is described in the experimental setting as an additional robustness evaluation.

\subsubsection{Context-Preserving Multimodal QA Construction}

The input pipeline adapts the hierarchical-question preprocessing of HiQuE~\cite{jung2024hique}. Each E-DAIC recording is transcribed into time-stamped segments with Whisper turbo~\cite{radford2023whisper,openai2024whisperturbo}. We then use the 85-question DAIC-WOZ inventory adopted by HiQuE, including each question's primary/follow-up designation, to recover the interviewer structure. A Sentence-Transformers \texttt{all-mpnet-base-v2} encoder embeds each candidate interviewer utterance and each inventory question; the normalized dot product selects the nearest inventory entry~\cite{reimers2019sentencebert,sentencetransformers2021mpnet}. Inventory aliases handle common ASR variants.

Question-table matching also repairs segments in which an interviewer question lacks a question mark or appears inside a longer ASR span. Matched question text is separated from adjacent participant text, with approximate timestamps assigned by character position within the original segment. Consecutive spans from the same speaker are merged. Files with unresolved turn boundaries are manually inspected, repaired, and resplit from the original waveform. Finally, every interviewer question is paired with all consecutive participant-response turns until the next question. This produces the context-preserving Multimodal QA Units consumed by the Layer~1 QA-to-Item Grounding Agent. Together with the PHQ-8 measurement contract described below, these procedures implement the non-agentic Stage~1 in Figure~2 of the main paper and precede all agent reasoning. Motivated by evidence that interviewer language can create predictive shortcuts~\cite{burdisso2024promptbias}, the procedure preserves interviewer questions as local context while treating participant responses as the primary basis for symptom assessment.

\subsubsection{Participant-Speech Acoustic Descriptor Alignment}

Only participant-response audio is analyzed. Recordings are resampled to 16~kHz and cut using the repaired transcript timestamps. The first branch applies openSMILE with \texttt{FeatureSet.eGeMAPSv02} and \texttt{FeatureLevel.Functionals}, yielding 88 eGeMAPS functionals per clip~\cite{eyben2010opensmile,eyben2016gemaps}. Each Multimodal QA Unit retains a compact subset covering mean pitch and pitch variability, mean loudness and loudness variability, local jitter, local shimmer, harmonic-to-noise ratio (HNR), and the first four MFCC means. The final text supplied to the Factor Specialists verbalizes pitch variability, loudness, jitter, shimmer, and HNR as coarse low/moderate/high descriptions. The bin boundaries are engineering quantization rules, not clinical cutoffs.

\begin{table}[H]
\centering
\scriptsize
\setlength{\tabcolsep}{3pt}
\begin{tabular}{@{}lll@{}}
\toprule
Descriptor & Low/moderate boundary & Moderate/high boundary \\
\midrule
Pitch variability & 0.15 & 0.45 \\
Mean loudness & 0.20 & 0.80 \\
Local jitter & 0.015 & 0.050 \\
Local shimmer (dB) & 0.40 & 1.20 \\
HNR (dB) & 5.0 & 20.0 \\
\bottomrule
\end{tabular}
\caption{Engineering thresholds used only to verbalize selected eGeMAPSv02 functionals. They do not define depression severity.}
\label{tab:appendix-egemaps-verbalization}
\end{table}

The second branch uses the official \texttt{KintsugiHealth/dam} Depression--Anxiety Model (DAM)~\cite{kintsugihealth2026dam}. DAM uses a fine-tuned Whisper-small.en acoustic backbone with task-specific depression and anxiety heads; its model card recommends at least 30 seconds of single-speaker English audio. Our preprocessing uses a 60-second minimum: consecutive participant-response clips are concatenated in interview order until a chunk contains at least 60 seconds of speech. A final shorter remainder is appended to the preceding chunk; if the participant has less than 60 seconds in total, DAM is skipped.

HiMA-MDD retains DAM-derived outputs as auxiliary acoustic descriptors. For the depression head, the quantized output is mapped to three bands: 0 for the model's PHQ-9 0--9 range, 1 for 10--14, and 2 for 15 or above. The resulting depression tag is attached to every participant answer contained in the corresponding chunk because individual answers are often too short for DAM inference. Participant-level summaries record the maximum label, duration-weighted mean label, and positive/severe chunk proportions. The DAM-derived descriptor supplies longer-context acoustic information, while the eGeMAPSv02 description supplies interpretable clip-level context. Both supplement the text and can be grounded in Layer~1 only to depressed mood, fatigue, concentration difficulty, and psychomotor disturbance; they remain auxiliary rather than decisive.

\subsubsection{PHQ-8 Measurement Contract and Targets}

The Patient Health Questionnaire-8 (PHQ-8) measures eight symptom categories over the preceding two weeks: anhedonia, depressed mood, sleep disturbance, fatigue, appetite disturbance, low self-worth, concentration difficulty, and psychomotor disturbance~\cite{kroenke2009phq8}. Each item takes an ordinal value in $\{0,1,2,3\}$, and the item sum ranges from 0 to 24. A systematic review and individual-participant-data meta-analysis reports comparable diagnostic accuracy for the PHQ-8 and PHQ-9 in screening settings~\cite{wu2020phq8phq9}.

In Figure~2 of the main paper, the PHQ-8 measurement contract refers to the combination of these item rubrics and ordinal semantics, the operational responsibility map used by the default Factor Specialists, and the fixed aggregation and threshold rules. The four responsibility groups are an implementation choice for evidence routing and provisional scoring, not a newly validated psychometric structure.

Every system predicts the complete eight-item profile. The item scores are summed deterministically, and totals of 10 or above are assigned to the depressed range for screening. The same item-scoring and screening rule is applied to gold and predicted profiles.

\subsection{Appendix B: Baselines and Evaluation Protocol}

\subsubsection{Baseline Adaptation}

The prompting baselines use the same Qwen2.5-72B-Instruct backbone and temperature-zero decoding as HiMA-MDD~\cite{qwen2024qwen25}. Zero-Shot directly predicts all eight PHQ-8 items, 3-Shot adds three labeled development examples, and Chain-of-Thought requests an explicit reasoning path before item prediction~\cite{wei2022cot}. Each baseline is run once per participant transcript.

MDAgents~\cite{kim2024mdagents} is adapted from its publicly released repository to receive a text-only PHQ-8 JSON task. AgentMental~\cite{hu2026agentmental} is likewise adapted from its publicly released repository, using its PHQ-8 topic and scoring resources, the completed interview transcript as simulated participant history, and the eight topic scores parsed from its final report. Its agents may retain the native follow-up flow, but each simulated answer is generated from that fixed history rather than obtained as a new observation from the original participant. The adaptation therefore evaluates AgentMental in a fixed-evidence setting rather than preserving the information-acquisition advantage of a real online interview. Both systems complete all 56 cases, and their outputs are normalized to the common eight-item schema. These adaptations align the publicly released repository implementations with the E-DAIC PHQ-8 evaluation protocol.

\subsubsection{Metrics}

Let $N$ denote the number of participants and $M=8$ the number of PHQ-8 items. For participant $n$ and item $i$, $y_{ni}$ and $\hat y_{ni}$ are the gold and predicted ordinal scores.

With $S_n=\sum_{i=1}^{M} y_{ni}$ and $\hat S_n=\sum_{i=1}^{M}\hat y_{ni}$, total-score errors are
\begin{align*}
\mathrm{Total\ MAE} &= \frac{1}{N}\sum_{n=1}^{N}|\hat S_n-S_n|,\\
\mathrm{Total\ RMSE} &= \sqrt{\frac{1}{N}\sum_{n=1}^{N}(\hat S_n-S_n)^2}.
\end{align*}

For screening, $c_n=\mathbb{I}[S_n\ge 10]$ and $\hat c_n=\mathbb{I}[\hat S_n\ge 10]$, where C and D denote the control-range and depressed-range classes. Accuracy is $N^{-1}\sum_{n=1}^{N}\mathbb{I}[c_n=\hat c_n]$. Cohen's $\kappa$ is
\begin{equation*}
\kappa=\frac{p_o-p_e}{1-p_e}, \qquad
p_e=\sum_{k\in\{\mathrm{C},\mathrm{D}\}}p_k\hat p_k,
\end{equation*}
where $p_o$ is observed agreement and $p_k$ and $\hat p_k$ are the gold and predicted proportions of class $k$. Item $\kappa$ applies the same unweighted definition to the four ordinal score categories for each item and then averages across the eight items.

For $k\in\{\mathrm{C},\mathrm{D}\}$, class-wise precision, recall, and F1 are
\begin{equation*}
\begin{aligned}
P_k &= \frac{\mathrm{TP}_k}{\mathrm{TP}_k+\mathrm{FP}_k},\\
R_k &= \frac{\mathrm{TP}_k}{\mathrm{TP}_k+\mathrm{FN}_k},\\
\mathrm{F1}[k] &= \frac{2P_kR_k}{P_k+R_k}.
\end{aligned}
\end{equation*}
We report $\mathrm{F1[C]}$, $\mathrm{F1[D]}$, and $\mathrm{Macro\text{-}F1}=(\mathrm{F1[C]}+\mathrm{F1[D]})/2$. A precision, recall, or F1 value with a zero denominator is set to zero. The primary evaluation reports Total MAE, Total RMSE, accuracy, screening $\kappa$, both class-wise F1 values, and Macro-F1. Table~1 reports MLlm-DR's published F1 value of 0.69 in the Macro-F1 column. Protocol-matched comparisons and significance tests use the locally rerun systems.

\subsubsection{Statistical Tests}

All star markers in the main comparison use the zero-shot LLM baseline as the sole reference. Total-score errors use one-tailed paired tests over shared test participants, accuracy uses the exact McNemar test, and class-wise F1 uses paired approximate randomization~\cite{dror2018significance}. These metric-specific tests are applied to aligned predictions. The resulting $p$-values are exploratory, nominal, and unadjusted for multiple comparisons. Each $p$-value corresponds to its reported metric. Screening $\kappa$ and Macro-F1 are outside the star-testing mechanism, and no marker denotes a comparison with MDAgents or AgentMental.

\subsubsection{Implementation and Reproducibility Details}

\begin{itemize}
    \item Primary test protocol: both gold and predicted PHQ-8 totals are thresholded at 10.
    \item Transcript ASR: Whisper turbo with time-stamped segments.
    \item Interview question inventory: 85 DAIC-WOZ questions with primary/follow-up types, adapted from HiQuE preprocessing.
    \item Inventory matching model: Sentence-Transformers \texttt{all-mpnet-base-v2} with normalized embeddings and cosine-equivalent dot product.
    \item Audio sample rate: 16~kHz mono.
    \item Hand-crafted acoustic representation: openSMILE eGeMAPSv02 functionals, with 88 extracted dimensions and a compact verbalized subset.
    \item Deep acoustic representation: \path{KintsugiHealth/dam}, checkpoint \texttt{dam3.1.ckpt}; 30-second non-overlapping internal windows and a 60-second minimum concatenated participant-response chunk in our wrapper.
    \item DAM use in HiMA-MDD: DAM-derived acoustic descriptors are retained as auxiliary input.
    \item Audio-eligible PHQ-8 items: depressed mood, fatigue, concentration difficulty, and psychomotor disturbance.
    \item Agent orchestration: LangGraph~\cite{langchain2026langgraph}.
    \item Runtime: Python 3.11.8, LangGraph 1.2.6, and scikit-learn 1.9.0.
    \item Backbone for HiMA-MDD and local comparisons: Qwen2.5-72B-Instruct.
    \item Decoding temperature: 0.
    \item Number of complete evaluation runs: one per method and configuration. A HiMA-MDD run contains multiple LLM calls across grounding, parallel Factor Specialists, global audit, any requested revision, and centralized reconstruction.
    \item Model serving and acceleration: vLLM~\cite{kwon2023vllm}.
    \item GPU platform: NVIDIA RTX 5880 Ada Generation with 48~GB memory.
    \item Maximum item-grounded QA records per PHQ-8 item: 8.
    \item Maximum deduplicated records per operational factor: 24.
    \item Maximum targeted specialist revision rounds: 1.
    \item Full post-hoc calibration feature dimension: 376.
    \item Post-hoc score-correction model comparison: standardized Ridge, Bayesian Ridge, and Elastic Net models, with hyperparameters selected on the development partition after fitting on the training partition.
    \item Reported post-hoc calibration model: eight independent Bayesian Ridge regressors fitted on the labeled training and development participants.
    \item Final post-hoc calibration fit: train and development partitions, 219 participants in total.
    \item Final post-hoc calibration application: the selected train+development models are applied to raw test predictions; test labels are used only for final evaluation.
    \item Corrected score conversion: nearest-integer rounding followed by clipping to $[0,3]$.
    \item Screening threshold: fixed at 10 and never tuned.
\end{itemize}

The evidence-index cache is keyed by the symptom schema, LLM-prompt version, model, temperature, and Multimodal-QA-Unit content hash. Cached candidate QA-to-item relations are shared across the controlled granularity runs so that the comparison changes the Factor-Specialist configuration and evidence access without rerunning Layer~1 grounding.

\begin{table*}[t!]
\subsection{Appendix C: Significance-Test Details}
\centering
\scriptsize
\setlength{\tabcolsep}{3pt}
\begin{tabular}{@{}llrrlrc@{}}
\toprule
System & Metric & System value & Zero-Shot value & Test & $p$ & Star \\
\midrule
HiMA-MDD (raw) & Total MAE & 3.9643 & 4.5357 & Paired $t$-test & 0.1347 & No \\
HiMA-MDD (raw) & Total RMSE & 4.9497 & 6.1296 & Paired $t$-test on squared error & 0.0161 & Yes \\
HiMA-MDD (raw) & Accuracy & 0.8036 & 0.7321 & Exact McNemar & 0.1719 & No \\
HiMA-MDD (raw) & F1[C] & 0.8493 & 0.8193 & Paired approximate randomization & 0.2113 & No \\
HiMA-MDD (raw) & F1[D] & 0.7179 & 0.4828 & Paired approximate randomization & 0.0154 & Yes \\
HiMA-MDD + Post-hoc Calibration & Total MAE & 3.4107 & 4.5357 & Paired $t$-test & 0.0122 & Yes \\
HiMA-MDD + Post-hoc Calibration & Total RMSE & 4.5728 & 6.1296 & Paired $t$-test on squared error & 0.0015 & Yes \\
HiMA-MDD + Post-hoc Calibration & Accuracy & 0.8393 & 0.7321 & Exact McNemar & 0.0352 & Yes \\
HiMA-MDD + Post-hoc Calibration & F1[C] & 0.8831 & 0.8193 & Paired approximate randomization & 0.0369 & Yes \\
HiMA-MDD + Post-hoc Calibration & F1[D] & 0.7429 & 0.4828 & Paired approximate randomization & 0.0120 & Yes \\
\bottomrule
\end{tabular}
\caption{Tests underlying the star markers in the main comparison. All tests use Zero-Shot as the reference and report metric-specific one-tailed $p$-values. The HiMA-MDD + Post-hoc Calibration condition uses the selected supervised Bayesian Ridge ordinal score-correction model. These exploratory values are nominal and unadjusted for multiple comparisons.}
\label{tab:appendix-significance}
\end{table*}

\end{document}